\documentclass[journal]{IEEEtran}
\include{inputs}
\usepackage{xcolor}

\usepackage{graphicx}
\graphicspath{{images/}}
\usepackage[section]{placeins}
\usepackage{enumitem,kantlipsum}
\usepackage{amsfonts}
\usepackage{amsmath}
\usepackage{caption}
\usepackage{subcaption}
\usepackage[linkcolor=blue,colorlinks=true]{hyperref}
\usepackage[ruled, vlined,linesnumbered ]{algorithm2e}
\usepackage{textcomp}
\DeclareMathOperator*{\argmin}{arg\,min}

\begin{document}
\title{Spatial Network Decomposition \\ for Fast and Scalable AC-OPF Learning}

\author{Minas Chatzos,
        Terrence~W.K.~Mak,~\IEEEmembership{Member,~IEEE,} 
        and~Pascal~Van~Hentenryck,~\IEEEmembership{Member,~IEEE}
\thanks{
The authors are affiliated with the 
School of Industrial and Systems Engineering,
Georgia Institute of Technology, Atlanta, GA 30332, USA. E-mail: minas@gatech.edu, wmak@gatech.edu, pvh@isye.gatech.edu.}
}

\markboth{}{}

\maketitle\sloppy\allowdisplaybreaks

\begin{abstract}
  This paper proposes a novel machine-learning approach for predicting
  AC-OPF solutions that features a fast and scalable training. It is
  motivated by the two critical considerations: (1) the fact that
  topology optimization and the stochasticity induced by renewable
  energy sources may lead to fundamentally different AC-OPF instances;
  and (2) the significant training time needed by existing
  machine-learning approaches for predicting AC-OPF. The proposed
  approach is a 2-stage methodology that exploits a spatial
  decomposition of the power network that is viewed as a set of
  regions. The first stage learns to predict the flows and voltages on
  the buses and lines coupling the regions, and the second stage
  trains, in parallel, the machine-learning models for each
  region. Experimental results on the French transmission system (up
  to 6,700 buses and 9,000 lines) demonstrate the potential of the
  approach. Within a short training time, the approach
  predicts AC-OPF solutions with very high fidelity and minor
  constraint violations, producing significant improvements over the
  state-of-the-art.  The results also show that the predictions can
  seed a load flow optimization to return a feasible solution within
  $0.03\%$ of the AC-OPF objective, while reducing running times
  significantly.
\end{abstract}


\IEEEpeerreviewmaketitle

\section{Introduction}
The \emph{AC Optimal Power Flow (AC-OPF)} problem is at the core of
modern power system operations. It determines the least-cost
generation dispatch that meets the demand of the power grid subject to
engineering and physical constraints. It is non-convex and NP-hard
\cite{7063278}, and the basic block of many applications, including
security-constrained OPF \cite{monticelli:87},
\cite{velloso2019exact}, security-constrained unit commitment
\cite{wang:08}, optimal transmission switching \cite{OTS}, capacitor
placement \cite{baran:89}, and expansion planning \cite{verma:16},
among others.

Machine learning has significant potential for real-time AC-OPF
applications for a variety of reasons \cite{chatzos2020highfidelity}.
A machine-learning model can leverage large amount of historical data
and deliver extremely fast approximations (compared to an AC-OPF
solver). Recent work (e.g.,
\cite{Fioretto:dnnopf,chatzos2020highfidelity}) has indeed shown that
machine-learning approaches can predict AC-OPF with high fidelity and
minimal constraint violations, using a combination of neural networks
and Lagrangian duality. However, the training times and memory
requirements of these machine-learning models can be quite
significant, which limits their potential applications. Indeed,
topology optimization and the stochasticity induced by renewable
energy sources may lead to fundamentally different AC-OPF instances
and it is unlikely that these approaches would scale to accommodate
the wide variety of input configurations encoutered in practice.
  
This paper explores a fundamentally different avenue: It seeks a
scalable machine-learning approach for predicting AC-OPF solutions
that can be trained quickly. Such an approach would make it possible
to train a machine-learning model quickly to accommodate a new
topology or a significant change in the capacity of renewable energy
sources. It would also open the possibility of training different
machine-learning models for different time periods.

To achieve this goal, the paper proposes a 2-stage machine-learning
approach that exploits a spatial decomposition of the power system.
The power network is viewed as a set of regions, the first stage
learns to predict the flows and voltages on the buses and lines
coupling the regions, and the second stage trains, in parallel, the
machine-learning models for each region. Experimental results on the
French transmission system (up to 6,700 buses and 9,000 lines)
demonstrate the potential of the approach. Within a short training
time, the approach predicts AC-OPF solutions with very high fidelity
and minor constraint violations, producing significant improvements
over the state-of-the-art.  The results also show that the predictions
can seed a load flow optimization to return a feasible solution within
$0.03\%$ of the AC-OPF objective, while reducing running times
significantly.
  
To our knowledge, the proposed approach is the first distributed
training algorithm for learning AC-OPF for large-scale network
topology. It builds on top, and significantly extends, prior work
\cite{Fioretto:dnnopf,chatzos2020highfidelity} combining machine
learning and Lagrangian duality. Most importantly, the 2-stage
approach significantly reduces the dimensionality of the learning
task, allows the training to be performed in parallel for each region,
and dramatically shortens training times, opening new avenues for
machine learning in very large-scale system operations. It is also the
first approach that can learn AC-OPF on an actual, large-scale
tranmission system fast, even on reasonable hardware configurations.

\section{Related work}

Machine learning has attracted significant attention in the power
systems community: recent overviews of the various approaches and
applications can be found in \cite{LouisSurvey,hasan20survey}.  In the
context of AC-OPF, various approaches have been proposed for learning
the active set of constraints \cite{misra2019learning},
\cite{xavier2019learning}, \cite{deka2019learning},
\cite{hasan2020hybrid}, \cite{robson2020learning}, imitating the
Newton-Raphson algorithm \cite{baker2020learningboosted}, or learning
warm-start points for speeding-up the optimization process
\cite{baker2019learning}, \cite{chen2020hotstarting}.  Several
approaches aim to predict optimal dispatch decisions
\cite{pan19deepopf}, \cite{pan2020deepopf},
\cite{DBLP:journals/corr/abs-1910-01213} but these were limited to
small-case studies. A load encoding scheme \cite{mak2021load} reduces
the dimensionality of AC-OPF instances, improving the results of
\cite{Fioretto:dnnopf,chatzos2020highfidelity}.  In the context of
DC-OPF, it is worth mentioning the results of
\cite{fd24f144965a4564b24e6062cbf936a5}, \cite{venzke2020learning}
that provide formal guarantees on the predictions of neural
networks. The application of machine learning to the
security-constrained extension of the DC-OPF is presented in
\cite{pan2020deepopf2}, \cite{velloso2020combining}.

\section{Preliminaries}

\paragraph{The AC Optimal Power Flow Problem}

A power network is modeled as an undirected graph $(\mathcal{N},
\mathcal{E})$ where $\mathcal{N}$ and $\mathcal{E}$ are the set of
buses and transmission lines. The set of generators and loads are
denoted by $\mathcal{G}$ and $\mathcal{L}$. The goal of the OPF is to
determine the generator dispatch of minimal cost that satisfies the
load. The OPF constraints include engineering and physical
constraints. The OPF formulation is shown in Figure \ref{fig:opf}.
The power flow equations are expressed in terms of complex power of
the form $S \!=\! (p \!+\! jq)$, where $p$ and $q$ denote the active and
reactive powers, admittances of the form $Y \!=\! (g \!+\! jb)$, where
$g$ and $b$ denote the conductance and susceptance, and voltages of
the form $V \!=\!  (v \angle \theta)$, with magnitude $v$ and phase
angle $\theta$.  The formulation uses $v_i, \theta_i, p_i^g$, and
$q_i^g$ to denote the voltage magnitude, phase angle, active power
generation, reactive power generation at bus $i$. Moreover, $p^f_{ij}$
and $q^f_{ij}$ denote the active and reactive power flows associated
with line $(i,j)$. The OPF receives as input the demand vectors
$p_i^d$ and $q_i^d$ for each bus $i$. The objective function captures
the cost of the generator dispatch. Typically, $c_i(\cdot)$ is a
linear or quadratic function. Constraints \eqref{con:2},
\eqref{con:3a}, and \eqref{con:3b} capture operating bounds for the
associated variables. The thermal limit for line $(i,j)$ is captured
via constraint \eqref{con:4}. Constraints \eqref{con:5a} and
\eqref{con:5b} capture \emph{Ohm's Law}. Constraints
\eqref{con:6a} and \eqref{con:6b} capture \emph{Kirchhoff's Law}.

\begin{figure}[!t]
\footnotesize
\begin{flalign*}
		& \textbf{minimize} \quad \sum_{i=1}^{\mathcal{|N|}} c_i(p_i^g) && 	\label{con:1} \tag{$1$}\\
		&\mbox{\bf subject to:} \notag\\
		&\hspace{6pt}
		\underline{v}_i \leq v_i \leq \overline{v}_i 		
		&& \!\!\!\!\!\forall i \in {\cal N} 		\label{con:2} \tag{$2$}\\
		&\hspace{6pt}
		\underline{p}_i^g \leq p_i^g \leq \overline{p}_i^g
		&& \!\!\!\!\!\forall i \in {\cal N} 		\label{con:3a} \tag{$3r$}\\
		&\hspace{6pt}
		\underline{q}_i^g \leq q_i^g \leq \overline{q}_i^g  
		&& \!\!\!\!\!\forall i \in {\cal N} 		\label{con:3b} \tag{$3i$}\\
		&\hspace{6pt}
		(p^f_{ij})^2 + (q^f_{ij})^2 \leq |\overline{S}|_{ij}^2	
		&& \!\!\!\!\!\forall (i,j) \in {\cal E}	\label{con:4}  \tag{$4$}\\
		&\hspace{6pt}
		 p_{ij}^f = g_{ij}v_i^2 -v_i v_j(b_{ij} \sin(\theta_i \text{-} \theta_j) + g_{ij}\cos(\theta_i \text{-} \theta_j))
		&& \!\!\!\!\!\forall (i,j) \in {\cal E} 	\label{con:5a} \tag{$5r$}\\
		&\hspace{6pt}
		q_{ij}^f = -b_{ij}v_i^2 - v_i v_j(g_{ij} \sin(\theta_i \text{-} \theta_j) - b_{ij}\cos(\theta_i \text{-} \theta_j))	
		&& \!\!\!\!\!\forall (i,j) \in {\cal E}		\label{con:5b} \tag{$5i$}\\
		&\hspace{6pt}
		p_i^g - p_i^d = \sum_{(i,j) \in \mathcal{E}} p_{ij}^f
		&& \!\!\!\!\!\forall i\in {\cal N} 		\label{con:6a} \tag{$6r$}\\
		&\hspace{6pt}
		q_i^g - q_i^d = \sum_{(i,j) \in \mathcal{E}} q_{ij}^f 
		&& \!\!\!\!\!\forall i\in {\cal N} 		\label{con:6b} \tag{$6i$}
\end{flalign*}
\caption{The OPF Formulation.}
\label{fig:opf}
\end{figure}

\paragraph{Neural Network Architectures}

Neural networks have achieved tremendous success in approximating
highly complex, nonlinear mappings in various domains and
applications. A Neural Network (NN) consists of a series of layers, the
output of each layer being the input to the next layer.  The NN layers 
are often fully connected and the function connecting
the layers is given by
$
\boldsymbol{o} = \pi(\boldsymbol{W} \boldsymbol{x} + \boldsymbol{b}),
$
where $\boldsymbol{x} \in \mathbb{R}^n$ is the input vector, $\boldsymbol{o} \!\in\!  \mathbb{R}^m$
the output vector, $\boldsymbol{W} \in \mathbb{R}^{m \times n}$ a weight matrix,
and $\boldsymbol{b} \in \mathbb{R}^m$ a bias vector. The function $\pi(\cdot)$ is
non-linear (e.g., a rectified linear unit (ReLU)).

\paragraph{Notations}

The cardinality of set $\mathcal{X}$ is denoted by
$|\mathcal{X}|$. $[N]$ represents the set $\{ 1,2, \dots, N
\}$. Vectors are displayed using bold letters and $\boldsymbol{x}$ $=
  [x_1, x_2, ..., x_n]^{\top}$. The element-wise lower (resp. upper)
  bound of the vector $\boldsymbol{x}$ is denoted by
  $\underline{\boldsymbol{x}}$ $(\mbox{resp. }
  \overline{\boldsymbol{x}})$. In learning algorithms, the prediction
  for $\boldsymbol{x}$ is denoted by $\hat{\boldsymbol{x}}$.

\section{Learning AC-OPF}

\subsection{OPF Learning Goals}

Given loads $(\boldsymbol{p^d}, \boldsymbol{q^d})$, the goal
is to predict the optimal setpoints $(\boldsymbol{p^g},
\boldsymbol{q^g})$ of the generators, the bus voltage
$\boldsymbol{v}$, and the phase angle difference $\boldsymbol{\Delta
  \theta}$ of the lines. This task is equivalent to learning the
nonlinear, high-dimensional mapping:
\begin{equation}
\mathcal{O}: \mathbb{R}^{2 |\mathcal{L}|} \to \mathbb{R}^{ \mathcal{|N|} + \mathcal{|E|} + 2 \mathcal{|G|}} \label{original}  \tag{$7$}
\end{equation}
\noindent
which maps the loads onto an optimal AC-OPF solution. The input to
the learning task is a dataset
$$\mathcal{D} = \{ (\boldsymbol{p^d}, \boldsymbol{q^d})^t,
(\boldsymbol{v}, \boldsymbol{\Delta \theta}, \boldsymbol{p^g},
\boldsymbol{q^g})^t \}_{t=1}^T$$ consisting of $T$ instances
specifying the inputs and outputs.

\subsection{A Lagrangian Dual Model for Learning AC-OPF}

One of the challenges of learning mapping $\mathcal{O}$ is the
presence of physical and engineering constraints. Ideally, given a 
NN $\mathcal{O}[\boldsymbol{w}]$ parameterized by weights
$\boldsymbol{w}$, the goal is to find the optimal solution
$\boldsymbol{w^*}$ of the problem:
\begin{align*}
 \min_{\boldsymbol{w}} & \quad \mathbb{L}_0(\boldsymbol{\hat{v}}, \boldsymbol{\hat{\theta}}, \boldsymbol{\hat{p}^g}, \boldsymbol{\hat{q}^g}) & \label{approxO} \tag{$10$} \\
\text{s.t.} & \quad (\boldsymbol{\hat{v}}, \boldsymbol{\hat{\theta}}, \boldsymbol{\hat{p}^g}, \boldsymbol{\hat{q}^g}) = \mathcal{O}[\boldsymbol{w}](\boldsymbol{p^d}, \boldsymbol{q^d}) \\
& \quad (\boldsymbol{\hat{v}}, \boldsymbol{ \hat{\theta}}, \boldsymbol{\hat{p}^g}, \boldsymbol{\hat{q}^g}, \boldsymbol{\hat{p}^f}, \boldsymbol{\hat{q}^f}) \text{ satisfy (\ref{con:2})-(\ref{con:6b}) }
\end{align*}
where $\mathbb{L}_0$ denotes the average norm of the difference
between the ground truth and the predictions
$\mathbb{L}_0(\boldsymbol{\hat{x}}) = \frac{1}{T} \sum_{t=1}^T \lvert
\lvert \boldsymbol{x}^t - \boldsymbol{\hat{x}}^t \rvert \rvert$ over
all training instances, and $(\boldsymbol{\hat{p}^f},
\boldsymbol{\hat{q}^f})$ are computed using constraints (\ref{con:5a})
and (\ref{con:5b}). However, it is unlikely that there exist weights
$\boldsymbol{w}$ such that the predictions actually satisfy the AC-OPF
constraints, since the learning task is a high-dimensional regression
task. However, ignoring the constraints entirely leads to predictions
that significantly violate the problem constraints as shown in
\cite{chatzos2020highfidelity,velloso2020combining}. The approach
from $\cite{Fioretto:dnnopf,chatzos2020highfidelity}$ addresses this
difficulty by using a Lagrangian dual method relying on constraint
violations.  The violation of a constraint $f(x) \geq 0$ is given by
$\nu_c(x) = \max \{0, -f(x) \}$, while the violation of $f(x) = 0$ is
$\nu_c(x) = |f(x)|$. Problem (\ref{approxO}) can then be approximated
by
\begin{align*}
 \min_{\boldsymbol{w}} & \quad \mathbb{L}(\boldsymbol{\lambda}, \boldsymbol{w}) = \mathbb{L}_0(\boldsymbol{\hat{v}}, \boldsymbol{\hat{\theta}}, \boldsymbol{\hat{p}^g}, \boldsymbol{\hat{q}^g}) + \boldsymbol{\lambda}^{\top} \bar{\boldsymbol{\nu}}  \label{approxdual} \tag{$11$} \\
\text{s.t.} & \quad (\boldsymbol{\hat{v}}, \boldsymbol{\hat{\theta}}, \boldsymbol{\hat{p}^g}, \boldsymbol{\hat{q}^g}) = \mathcal{O}[\boldsymbol{w}](\boldsymbol{p^d}, \boldsymbol{q^d})
\end{align*}
where $\boldsymbol{\lambda}^{\top} \bar{\boldsymbol{\nu}} = \sum_{c
  \in \mathbb{C}} \lambda_c \bar{\nu}_c(\boldsymbol{\hat{v}},
\boldsymbol{\hat{\theta}}, \boldsymbol{\hat{p}^g},
\boldsymbol{\hat{q}^g})$, $\lambda_c$ is the weight for the violation
of constraint $c$, and $\bar{\nu}_c$ denotes the average violation of
constraint $c$ over all training instances. Again, the satisfaction of
the constraints (\ref{con:5a}), (\ref{con:5b}) is guaranteed, since
the power flows are computed indirectly from these constraints. For a
fixed $\boldsymbol{\lambda}$, $\mathbb{L}(\boldsymbol{\lambda},
\boldsymbol{w})$ can be used as the loss function for training the
neural network.  Moreover, the constraint weights can be updated using
a subgradient method that performs the following operations in
iteration $j$.
\begin{align*}
    \boldsymbol{w}^j = \argmin_{\boldsymbol{w}} \mathbb{L}(\boldsymbol{\lambda}^{(j-1)}, \boldsymbol{w}) \tag{$12$} \label{originalalg}\\
    \boldsymbol{\lambda}^j = \boldsymbol{\lambda}^{(j-1)} + \rho \boldsymbol{\bar{\nu}}(\boldsymbol{w}^{j})
\end{align*}

Learning the mapping $\mathcal{O}$ is challenging for large-scale
topologies. For instance, it takes 7 hours to train a network for a
topology of 3500 buses \cite{chatzos2020highfidelity}. This limits the
potential applications of neural networks in large power systems which
may be up to 50000 buses.  Indeed, during operations, the topology of
the system may change from day to day through line or bus switching,
meaning that a different mapping needs to be learned. Similarly, the
mapping $\mathcal{O}$ depends on the commitment decisions in the
day-ahead markets, again potentially changing the mapping to be
learned.


The goal of this paper is to propose {\em a fast training procedure to
  learn the mapping $\mathcal{O}$}. Such a fast training procedure
would have many advantages: the NN model could be trained after the
day-ahead market clearing and/or in real time during operations when
the network topology changes, and it could be tailored to the load
profiles of specific times in the day (e.g., 2:00pm-4:00pm).  These
considerations are important, especially given the increasing share of
renewable energy in the energy mix and the increasing prediction
errors.

\subsection{Exploiting Network Sparsity}

One possible avenue to obtain a fast training procedure is to exploit
the sparsity typically found in power system networks. Consider a
partition $\{ \mathcal{N}^{k} \}_{k=1}^K$ of the buses, i.e.,
$$
\bigcup_{k=1}^K  \mathcal{N}^{k} = \mathcal{N}, \quad \mathcal{N}^{k} \cap \mathcal{N}^{k'} = \emptyset, k \neq k'
$$
Denote the generators and loads of region $k$ by $\mathcal{G}^k$ and $\mathcal{L}^k$ respectively and define 
$$
\mathcal{E}^k = \{ (i, j) \in \mathcal{E} : i,j \in \mathcal{N}^k \}, k \in [K].
$$
$$
\mathcal{E}^{\leftrightarrow} = \mathcal{E} \setminus ( \cup_{k=1}^K \mathcal{E}^k), \quad \mathcal{N}^{\leftrightarrow} = \{ i: (i,j) \in \mathcal{E}^{\leftrightarrow} \ \vee (j,i) \in \mathcal{E}^{\leftrightarrow} \}
$$ Here $\mathcal{E}^k$ represents the lines within partition element
$k$ and $\mathcal{E}^{\leftrightarrow}$ the coupling lines that
connect partition elements. In the French transmission system, the
test case in this paper, $|\mathcal{N}| = 6705$, $|\mathcal{E}| =
8962$, and $|\mathcal{E}| \approx 1.3 \times |\mathcal{N}|$. Moreover,
the system is organized in $12$ geographical areas using $326$
$(3.6\%)$ coupling lines and $\max_{k \in [K]} |\mathcal{N}^k| = 1156$
$(17.2\%$) buses.

To leverage the network sparsity, a natural first attempt would be to
learn a mapping for each region, i.e.,
\begin{equation}
\mathcal{O}_0^k: \mathbb{R}^{2|\mathcal{L}^k|} \to \mathbb{R}^{2|\mathcal{N}^k| + 2|\mathcal{G}^k|} \quad (k \in [K]). \label{regionalmap0}  \tag{$8$}
\end{equation}
The learning thus predicts the setpoints for generators in region
$k$ using only the loads of the same region. These learning tasks
would be performed independently and in parallel. However, it is
obvious that the loads $\mathcal{L}^k$ are not sufficient to determine
the optimal setpoints for generators $\mathcal{G}^k$. In fact,
$\mathcal{O}_0^k$ is not even a function, since two inputs for the
loads $\mathcal{L}^k$ in the training set may be associated with
different outputs due to loads in other parts of the network.

\subsection{Capturing Flows on Coupling Lines}

\begin{figure}[t]
\centering
\includegraphics[width=0.5\textwidth,scale=0.8]{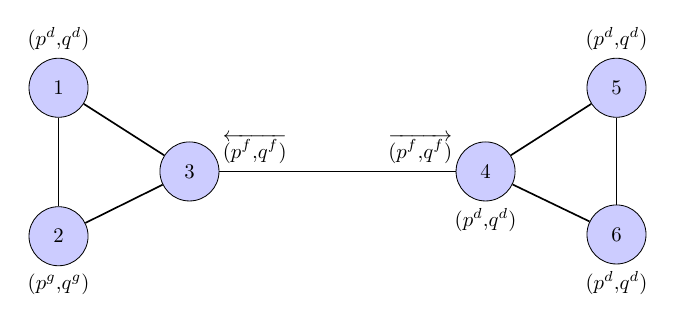}
\caption{A Simple Network With $|\mathcal{N}|=6$, $|\mathcal{E}|=7$, $|\mathcal{G}|=1$, and $|\mathcal{L}|= 4$.}
\label{fig:simplePS}
\end{figure}

Consider the simplistic power system depicted in Figure
\ref{fig:simplePS}. There are two areas, $\mathcal{N}^1 = \{1,2,3 \}$
and $\mathcal{N}^2 = \{4,5,6 \}$, which gives
$\mathcal{N}^{\leftrightarrow} = \{ 3,4 \}$,
$\mathcal{E}^{\leftrightarrow} = \{(3,4), (4,3) \}$. The mapping
$\mathcal{O}$ views the setpoints for the generator at bus 2 $(p^g_2,
q^g_2)$ as a function of $(p^d_1, q^d_1, p^d_4, q_4^d, p_5^d, q_5^d,
p_6^d, q_6^d)$. However, assume that flows $(p_{4,3}^f, q_{4,3}^f)$,
along with the voltage magnitudes $v_3, v_4$ are fixed and respect
constraints (\ref{con:4}), (\ref{con:5a}), (\ref{con:5b}) associated
with line $(3,4)$. In that case, the setpoints for the generator at bus 2 can
be computed without the knowledge of $(p^d_4, q_4^d, p_5^d, q_5^d,
p_6^d, q_6^d)$: the vector $(p_{4,3}^f, q_{4,3}^f, v_3)$ encodes all
the information from area 2 needed to compute the generator setpoint.
Hence, one may attempt to express $(p^g_2, q^g_2)$ as a function of
$(p^d_1, q^d_1, p_{4,3}^f, q_{4,3}^f, v_3)$ which decreases the input
size from 8 to 5. The input size decreases by 3 in this example but
the size reduction is significantly larger in actual systems.

With this in mind, the mappings in Equation \ref{regionalmap0} 
become
\begin{equation}
\mathcal{O}^k: \mathbb{R}^{2|\mathcal{L}^k| + |\mathcal{N}^{\to k}|+ 2|\mathcal{E}^{\to k}|} \to \mathbb{R}^{|\mathcal{N}^k \setminus \mathcal{N}^{\to k}| + |\mathcal{E}^k| + 2|\mathcal{G}^k|} \label{regionalmap}  \tag{$9$}
\end{equation}
where the coupling lines, buses of region $k$ are defined as
\begin{align*}
\mathcal{E}^{\to k} = \{(i,j) \in \mathcal{E}^{\leftrightarrow}: i \in \mathcal{N}^k \ \vee j \in \mathcal{N}^k \} \\
\mathcal{N}^{\to k} = \mathcal{N}^k \cap \mathcal{N}^{\leftrightarrow}
\end{align*}
$\mathcal{O}^k$ maps the loads in area $k$, the flows to area $k$, and
the voltage of the coupling buses to the optimal generator setpoints
in the area, i.e., the active and reactive outputs of the regional
generators, the phase angle differences of the regional branches, and
the voltage setpoints for the non-coupling buses of the region. For
large transmission systems, the input/output dimensions of each
mapping $\mathcal{O}^k$ are significantly smaller that those of
$\mathcal{O}$.  The learning tasks can proceed in parallel and their
complexity is reduced, since each mappings $\mathcal{O}^k$ is an order
of magnitude smaller in size than $\mathcal{O}$.

Unfortunately, this approach has a key limitation: {\em each mapping
  $\mathcal{O}^k$ can be learned from historical data but cannot be
  used for prediction since the coupling flows and voltages are not
  known at prediction time}. Indeed, during training, the learning
task has access to the coupling values for each instance. However,
this is not true at prediction time. The next section shows how to
overcome this difficulty.

\section{Two-Stage Learning of AC-OPF}

The fast training method for AC-OPF is a two-stage approach: the first
stage is a NN that predicts the flow on the coupling lines and the
second stage is a collection of NNs, each of which approximates a
mapping $\mathcal{O}^k$.

\subsection{Learning Coupling Voltages \& Flows}

The goal of the first stage is to learn the mapping
\begin{equation}
    \mathcal{O}^{\leftrightarrow}:\mathbb{R}^{2|\mathcal{L}|} \to \mathbb{R}^{|\mathcal{N}^{\leftrightarrow}| + |\mathcal{E}^{\leftrightarrow}|}  \label{couplingmap}  \tag{$13$}
\end{equation}
\noindent from the loads to the voltages magnitude $\boldsymbol{v}^0$
of the coupling buses and the phase angle difference $\boldsymbol{\Delta
  \theta}^0$ of the coupling branches. Although the mapping considers
all loads, it can be learned fast (e.g., under 30 minutes) even for
large networks, because of the small number of coupling buses and
lines. The coupling flows are then computed indirectly via constraints
(\ref{con:5a}) and (\ref{con:5b}). Let $\mathbb{C}^{\leftrightarrow}$
denote the set of constraints (\ref{con:2}) for $i \in
\mathcal{N}^{\leftrightarrow}$, and (\ref{con:4}) and (\ref{con:5a}),
(\ref{con:5b}) for $(i,j) \in \mathcal{E}^{\leftrightarrow}$.  The
learning task uses a neural network
$\mathcal{O}^{\leftrightarrow}[\boldsymbol{w}^0]$ parameterized by
weights $\boldsymbol{w}^0$ and predicts the coupling voltages
$(\boldsymbol{\hat{v}}^0, \boldsymbol{\widehat{\Delta \theta}}^0)$.
The loss function is given by:
\[
\mathbb{L}^{\leftrightarrow}(\boldsymbol{\lambda}, \boldsymbol{w}^0) = \mathbb{L}_0(\boldsymbol{\hat{v}}^0, \boldsymbol{\widehat{\Delta \theta}}^0) + \sum_{c \in \mathbb{C}^{\leftrightarrow}} \lambda_c \bar{\nu}_c(\boldsymbol{\hat{v}}^0, \boldsymbol{\widehat{\Delta \theta}}^0)
\]
The training follows equation (\ref{originalalg}) and the resulting optimal weights $(\boldsymbol{w}^0)^*$ lead to the first-stage predictions
$$
(\boldsymbol{\hat{v}}^0, \boldsymbol{\widehat{\Delta \theta}}^0) = \mathcal{O}^{\leftrightarrow}[(\boldsymbol{w}^0)^*](\boldsymbol{p^d}, \boldsymbol{q^d})
$$ and the resulting first-stage coupling flow predictions
$(\boldsymbol{\hat{p}^f})^0, (\boldsymbol{\hat{q}^f})^0$. The first
stage is summarized in Algorithm 1.

\begin{algorithm}[!t]
\footnotesize
\caption{The First-Stage Coupling Training.}
\SetAlgoLined
        $\lambda_c \leftarrow 0, \forall c \in \mathbb{C}^{\leftrightarrow}$\\
        \For{$i = 1,2,..., epochs_{\lambda}$}{
            \For{$j = 1,2,..., epochs_{w}$}{
                $(\boldsymbol{\hat{v}}^0, \boldsymbol{\widehat{\Delta \theta}}^0) \leftarrow \mathcal{O}^{\leftrightarrow}[\boldsymbol{w}^0] (\boldsymbol{p^d}, \boldsymbol{q^d})$ \\
                $\mathbb{L}^{\leftrightarrow}(\boldsymbol{\lambda}, \boldsymbol{w}^0) \leftarrow \mathbb{L}_0(\boldsymbol{\hat{v}}^0, \boldsymbol{\widehat{\Delta \theta}}^0) + \sum_{c \in \mathbb{C}^{\leftrightarrow}} \lambda_c \bar{\nu}_c(\boldsymbol{\hat{v}}^0, \boldsymbol{\widehat{\Delta \theta}}^0)$ \\
                $\boldsymbol{w}^0 \leftarrow \boldsymbol{w}^0 - \alpha \nabla_{\boldsymbol{w}^0} (\mathbb{L}^{\leftrightarrow}(\boldsymbol{\lambda}, \boldsymbol{w}^0))$
                }
                $\lambda_c \leftarrow \lambda_c + \rho \bar{\nu}_c, \quad \forall c \in \mathbb{C}^{\leftrightarrow}$
  }
\KwResult{Weights $(\boldsymbol{w}^0)^*$}
\end{algorithm}

\subsection{Training of Regional Systems}

\begin{figure}[!t]
\centering
\includegraphics[width=0.5\textwidth,scale=0.8]{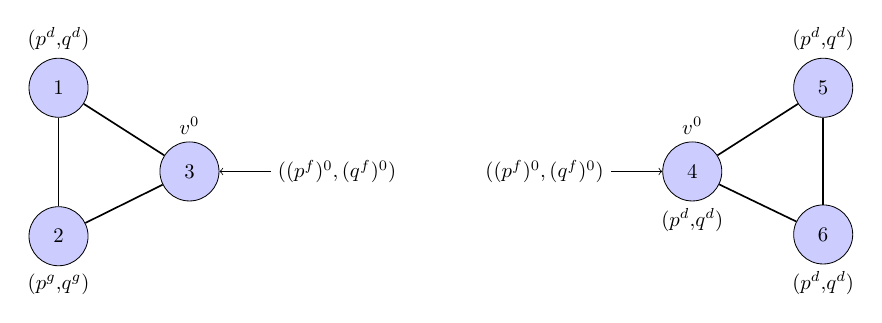}
\caption{Illustration of the Decomposition.}
\label{fig:simpledecomp}
\end{figure}

The training of the regional systems uses the first-stage predictions
for the coupling flows and voltages. Note however that it could use
the ground truth present in the instance data, but experimental
results have shown that this degrades the overall prediction accuracy.
The decoupling is illustrated in Figure \ref{fig:simpledecomp}, where
the voltages of the coupling buses $3$ and $4$ and the
incoming/outgoing flows for each region are fixed to the first-stage
predictions.

To learn mappings $\mathcal{O}^k \ (k \in [K])$, let $\mathbb{C}^k$
denote the set of constraints associated with region $k$, i.e.,
constraint (\ref{con:2}) for buses $i \in \mathcal{N}^k \setminus
\mathcal{N}^{\to k}$, constraints (\ref{con:3a}), (\ref{con:3b}),
(\ref{con:6a}), and (\ref{con:6b}) for buses $i \in \mathcal{N}^k$,
and constraints (\ref{con:4}), (\ref{con:5a}), (\ref{con:5b}) for
branches $(i,j) \in \mathcal{E}^k$.  In particular, the power balance
constraint (\ref{con:6a}), (\ref{con:6b}) for region $k$ becomes
\begin{align*}
&p_i^g - (p_i^d -\sum_{(i,j) \in \mathcal{E}^{\to k}} (\hat{p}_{ij}^f)^0) = \sum_{(i,j) \in \mathcal{E}^k} p_{ij}^f  & i \in \mathcal{N}^k \\
&q_i^g - (q_i^d - \sum_{(i,j) \in \mathcal{E}^{\to k}} (\hat{q}_{ij}^f)^0)  = \sum_{(i,j) \in \mathcal{E}^k} q_{ij}^f & i \in \mathcal{N}^k
\end{align*}
The learning task uses a collection $\{\mathcal{O}^k[\boldsymbol{w}^k] \}_{k \in [K]}$ of NNs and the loss function for each regional net is given by
\begin{align*}
\mathbb{L}^{k}(\boldsymbol{\lambda}, \boldsymbol{w}^k) = \mathbb{L}_0(\boldsymbol{\hat{v}}^k, \boldsymbol{\widehat{\Delta \theta}}^k, (\boldsymbol{\hat{p}^g})^k, (\boldsymbol{\hat{q}^g})^k) + \\
\sum_{c \in \mathbb{C}^{k}} \lambda_c \bar{\nu}_c(\boldsymbol{\hat{v}}^k, \boldsymbol{\widehat{\Delta \theta}}^k, (\boldsymbol{\hat{p}^g})^k, (\boldsymbol{\hat{q}^g})^k, \boldsymbol{\hat{v}}^{0,k}, (\boldsymbol{\hat{p}^f})^{0,k}, (\boldsymbol{\hat{q}^f})^{0,k})
\end{align*}
where $\boldsymbol{\hat{v}}^{0,k}$ is the first-stage prediction for
the voltage magnitude of the coupling buses of region $k$ and
$(\boldsymbol{\hat{p}^f})^{0,k}, (\boldsymbol{\hat{q}^f})^{0,k}$ the
first-stage predictions for the incoming/outgoing flows of region
$k$. The training, summarized in Algorithm 2, is performed using the
approach in equation (\ref{originalalg}) and each region can be
trained in parallel.  Line 2 predicts the voltage setpoints for the
coupling buses and the phase angle differences of the coupling
lines. Line 3 computes the predicted coupling flows from these
predictions. Line 6 computes the predictions for region $k$ given the
current NN parameters and constraint weights. Line 8 performs the
back-propagation to update the weights and line 10 updates the
constraint weights.

\begin{algorithm}[!tb]
\footnotesize
\caption{The Second-Stage Training (Region $k$).}
\SetAlgoLined
$\lambda_c \leftarrow 0, \forall c \in \mathbb{C}^k$ \\
                $(\boldsymbol{\hat{v}}^{0}, \boldsymbol{\widehat{\Delta \theta}}^{0}) \leftarrow \mathcal{O}^{\leftrightarrow}[(\boldsymbol{w}^0)^*] (\boldsymbol{p^d}, \boldsymbol{q^d})$ \\
                \text{Compute} $(\boldsymbol{\hat{p}^f})^{0,k}, (\boldsymbol{\hat{q}^f})^{0,k}$ \text{via (\ref{con:5a}), (\ref{con:5b})} \\

        \For{$i = 1,2,..., epochs_{\lambda}$}{
            \For{$j = 1,2,..., epochs_{w}$}{
                $(\boldsymbol{\hat{v}}^k, \boldsymbol{\widehat{\Delta \theta}}^k, (\boldsymbol{\hat{p}^g})^k, (\boldsymbol{\hat{q}^g})^k) \leftarrow \mathcal{O}^{k}[\boldsymbol{w}^k] ((\boldsymbol{p^d})^k, (\boldsymbol{q^d})^k, \boldsymbol{\hat{v}}^{0,k}, (\boldsymbol{\hat{p}^f})^{0,k}, (\boldsymbol{\hat{q}^f})^{0,k})$ \\
                $\mathbb{L}^{k}(\boldsymbol{\lambda}, \boldsymbol{w}^k) \leftarrow \mathbb{L}_0(\boldsymbol{\hat{v}}^k, \boldsymbol{\widehat{\Delta \theta}}^k, (\boldsymbol{\hat{p}^g})^k, (\boldsymbol{\hat{q}^g})^k) + \sum_{c \in \mathbb{C}^{k}} \lambda_c \bar{\nu}_c((\boldsymbol{\hat{v}}, \boldsymbol{\widehat{\Delta \theta}}, \boldsymbol{\hat{p}^g}, \boldsymbol{\hat{q}^g}, \boldsymbol{\hat{v}}^{0}, (\boldsymbol{\hat{p}^f})^{0}, (\boldsymbol{\hat{q}^f})^{0})^k)$\\
                $\boldsymbol{w}^k \leftarrow \boldsymbol{w}^k - \alpha \nabla_{\boldsymbol{w}^k} (\mathbb{L}^{k}(\boldsymbol{\lambda}, \boldsymbol{w}^k))$
                } 
                $\lambda_c \leftarrow \lambda_c + \rho \bar{\nu}_c, \quad \forall c \in \mathbb{C}^{k}$
                }
\KwResult{Weights $(\boldsymbol{w}^k)^*$}
\end{algorithm}

\newpage
\normalsize
\section{Experimental Results}

\subsection{Experimental Setting}

The datasets were generated by solving AC-OPF problems with the
nonlinear solver IPOPT \cite{wachter06on} using 2.5 GHz-i7 Intel Cores
and 16GB of RAM. In total, $10^4$ load profiles, which correspond to
feasible AC-OPF problems, were generated for each test case. $80\%$ of
these instances were used for training and the remaining $20\%$ for
testing. The learning models were implemented using PyTorch
\cite{paszke2017automatic} and trained using NVidia Tesla V100 GPUs
with 16GB of memory. The training of each network utilizes
mini-batches of size 120 and the learning rate $\alpha$ was set to be
decreasing from $10^{-3}$ to $10^{-6}$, while $\rho$ was set to
$10^{-3}$.

The first-stage NN consists of two subnetworks with sizes
$2|\mathcal{L}| \times |\mathcal{N}^{\leftrightarrow}| \times
|\mathcal{N}^{\leftrightarrow}|$ and $2|\mathcal{L}| \times
|\mathcal{E}^{\leftrightarrow}| \times
|\mathcal{E}^{\leftrightarrow}|$ for the voltage magnitudes and phase
angle differences respectively.  For each region, each NN topology
consists of 4 subnetworks, one for each predicted variable. The
subnetworks have one hidden layer of size $ 2|\mathcal{L}^k| \times
3|\mathcal{L}^k| \times |\mathcal{G}^k|$. The subnetworks used to
approximate the original mapping (equation (\ref{original})) are
similar in structure and have size $ 2|\mathcal{L}| \times
3|\mathcal{L}| \times |\mathcal{G}|$.

\subsection{Load Profiles}

\begin{table*}[!t]
\caption{The Power System Networks with Regional Information}
    \begin{center}
    \def\arraystretch{1.2}%
        \begin{tabular}{|c |c c c c| c c|} 
         \hline
        Benchmark & $|\mathcal{N}|$ & $|\mathcal{E}|$ & $|\mathcal{L}|$ & $|\mathcal{G}|$ & $|\mathcal{N}^k|_{k=1}^K$ & $|\mathcal{E}^{\leftrightarrow}|$\\ [0.5ex] 
        \hline\hline
        {\tt France\_EHV} & 1737 & 2350 & 1731 & 290  & [338, 280, 233, 179, 143, 126, 124, 72, 67, 64, 57, 54] & 148\\ 
        \hline
        {\tt France\_Lyon} & 3411 & 4499 & 3273 & 771 & [1158, 357, 294, 288, 264, 255, 231, 197, 184, 67, 62, 54] & 219\\
        \hline
        {\tt France} & 6705 & 8962 & 6262 & 1708 & [1156, 796, 748, 746, 627, 517, 497, 395, 325, 322, 298, 278] & 326\\
        \hline
        \end{tabular}
\end{center}
\label{table:networks}
\end{table*}

The power systems used as test cases (Table \ref{table:networks}) are
parts of the actual French Transmission System. {\tt France} is the
French transmission system, {\tt France\_EHV} is the very high-voltage
French system, and {\tt France\_LYON} is {\tt France\_EHV} with a
detailed representation of the Lyon region. The French system is
organized in 12 geographical regions. The dataset is generated by
taking into account this geographical information. A load $l$ in
region $k$ with nominal value $(p^d, q^d)^0$ is generated with the
formula
$$
(p^d, q^d) = (\alpha + \beta^k + \gamma^l) ((p^d)^0, (q^d)^0)
$$ where the following coefficients are randomly drawn from the
folowing distributions
\begin{align*}
    \alpha \sim \text{Uniform}[0.875, 0.975] \\
    \beta^{k} \sim \text{Uniform}[-0.025, 0.025], \quad \forall k \in [K] \\
    \gamma^{l} \sim \text{Uniform}[-0.0025, 0.0025], \quad \forall l \in \mathcal{L}
\end{align*}
The term $\alpha$ captures the system-wide load level, while $\beta^k$
is associated with differences in the loads between regions (e.g., due
to potentially different weather conditions). The difference in
coefficients may be up to $5\%$ for two different regions. Finally,
$\gamma^l$ is the uncorrelated noise added to each individual load
with a range of $0.5\%$ of its nominal value.

\begin{figure}[!t]
\centering
\includegraphics[width=0.5\textwidth,scale=0.56]{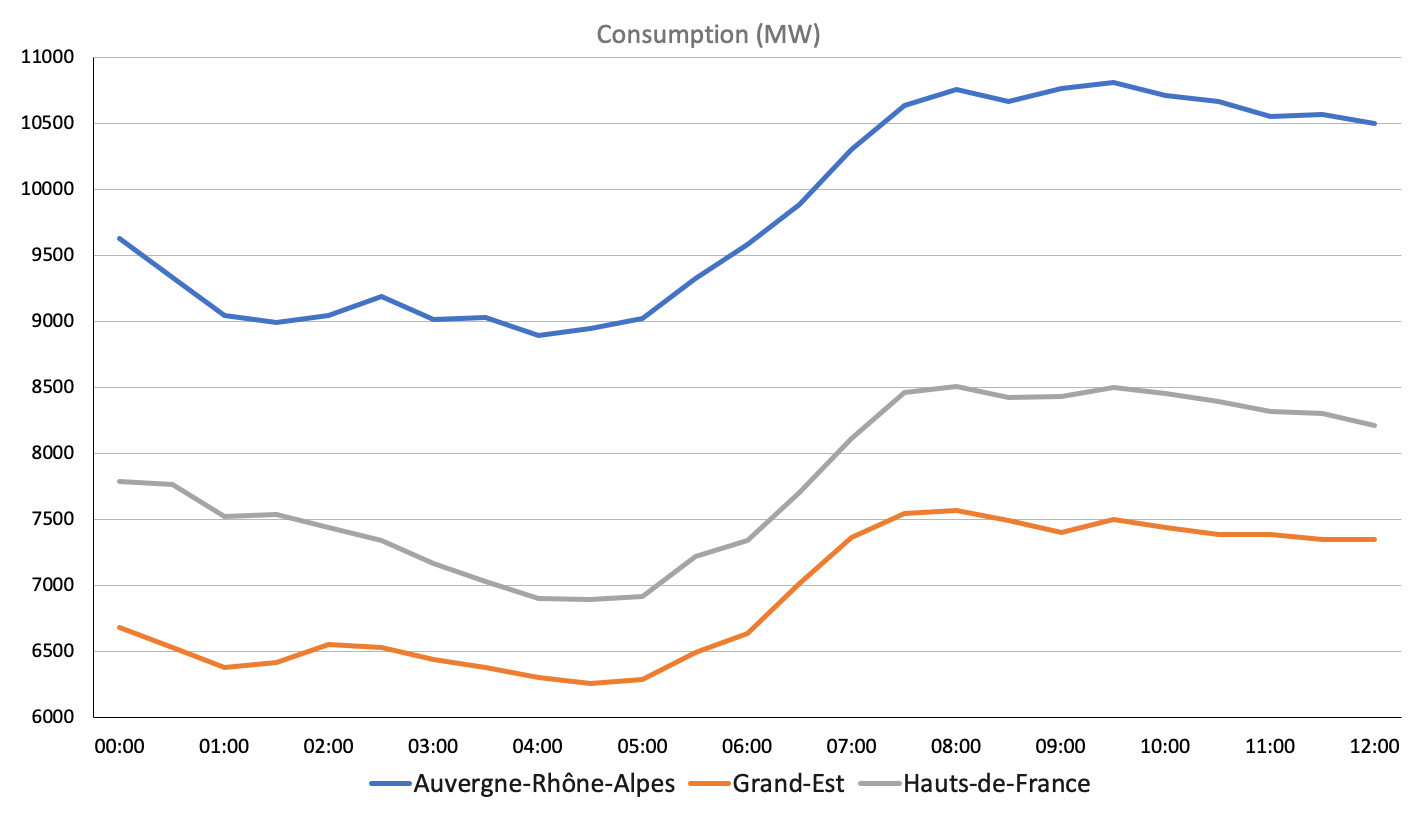}
\caption{Consumption for Three Regions in the French System over a 12-hour Interval.}
\label{fig:consuption}
\end{figure}

The resulting dataset captures realistic load profiles: the uniform
load perturbation, the load level differences between the regions, and
the fixed active/reactive power ratio represent the typical behavior
for aggregated demand in a large-scale topology spanning several
geographic regions. Randomly perturbing each individual load in an
uncorrelated fashion would produce unrealistic load profiles: they
would lead to an unnecessarily challenging learning task that would
need to capture an exponential number of unrealistic behaviors of the
power system. To highlight this point, Figure
\ref{fig:consuption} depicts the actual consumption for three French
regions over a 12 hour interval. Observe the strong correlation of the
demand between the three regions. However, the correlation is not
perfect and the ratios between the regional loads vary by small
factors. The term $\beta_k$ is used to account for this behavior. The
resulting load profiles range from 0.85 of the nominal to the nominal
load. This $15\%$ difference is typical over a 12-hour interval as
shown in Figure \ref{fig:consuption}.

\subsection{First-stage Predictions}

\begin{figure}[!t]
\centering
\includegraphics[scale=0.24]{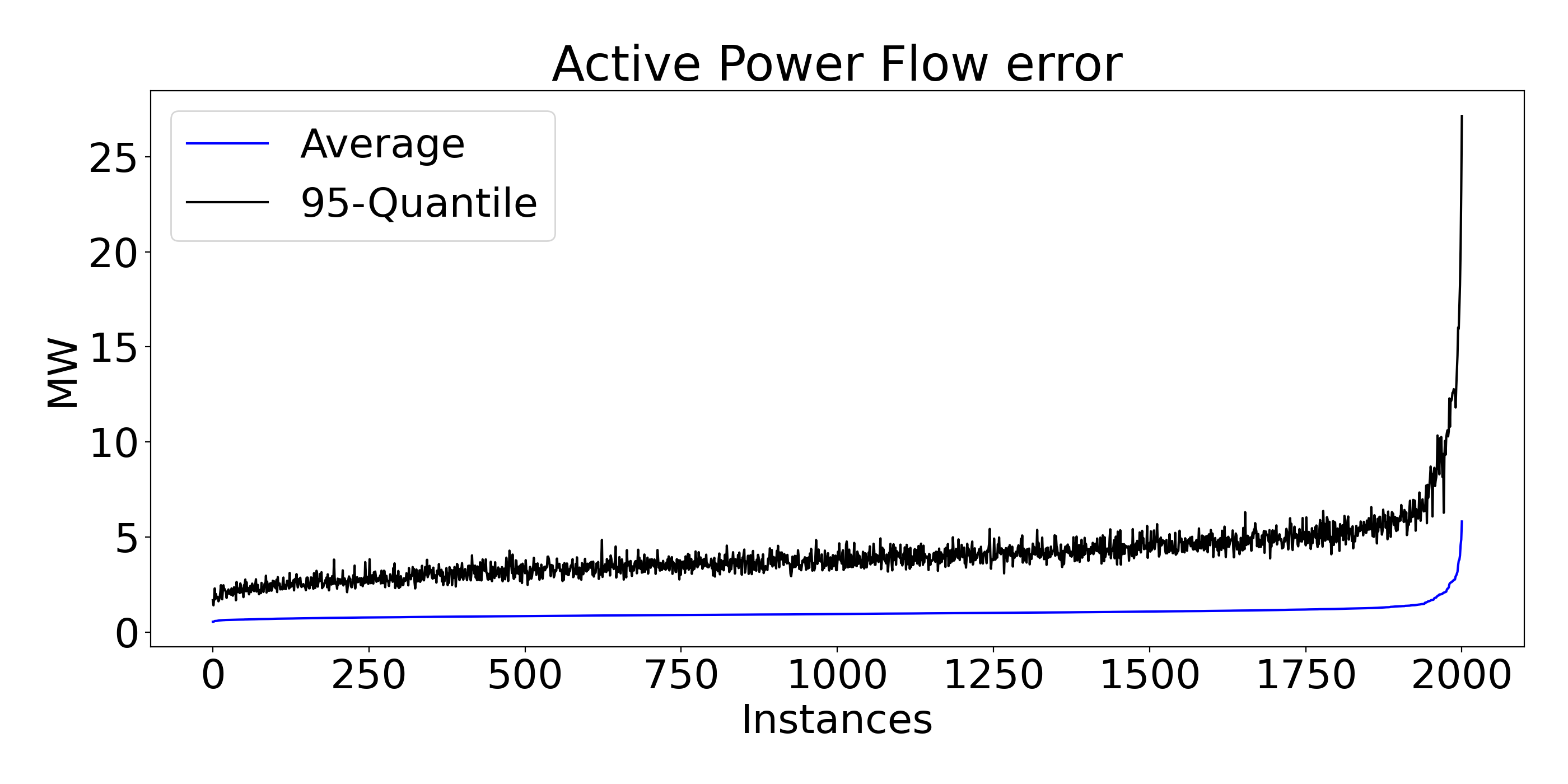}\\
\caption{Prediction Errors (Average and 95$\%$ Quantile) over all Testing Instances for the Active Flow of the Coupling Branches for Testcase {\tt France}. The Instances are Sorted in Increasing Order of Average Error.}
\label{fig:firststage}
\end{figure}

This section presents the prediction errors of the first stage. The
training time was limited to 30 minutes. Table \ref{table:cferrors}
contains aggregate results for the active and reactive powers of the
coupling branches, as well the voltage magnitudes, for all three test
cases. The results are an average over all instances and coupling
branches. The average error is close to $1 \ \mbox{MW}$ for the
largest two test cases: {\tt France} and {\tt
  France\_Lyon}. Meanwhile, the $95$-Quantile indicates that $95$\% of
the predictions result in an error less that $5 \ \mbox{MW}$. In the
smaller {\tt France\_EHV}, the errors are slightly higher reaching
$3.5 \ \mbox{MW}$ on average. Given that the nominal load of the {\tt
  France} system is $50,000 \ \mbox{MW}$) and the nominal flow values
for most coupling branches are greater than $100 \ \mbox{MW}$ and go up
to $1,000 \ \mbox{MW}$), these results indicate that the prediction
errors are small in percentage for all test cases. Table
\ref{table:cferrors} also shows that the voltage magnitudes are
predicted very accurately. Figure \ref{fig:firststage} contains
detailed results on the active part of the flows of the coupling
branches for the {\tt France} test case, showing consistent results
across all tested instances. The $95\%$ quantile graph indicates that
the prediction errors exceed $5 \ \mbox{MW}$ only for a very small
percentage of the test cases and branches.

\begin{table}[!t]
    \begin{center}
    \def\arraystretch{1.2}%
        \begin{tabular}{|c|c c c c|} 
        \hline
        & \multicolumn{2}{|c|}{$\hat{p}^f$ (MW)} & \multicolumn{2}{|c|}{$v$ (P.U)} \\
        \hline
        Benchmark & Avg & $95\%$ Quantile & Avg & $95\%$ Quantile\\ [0.5ex] 
        \hline\hline
        {\tt France\_EHV} & 3.43 & 11.91 & 25 $\cdot 10^{-5}$ & $76 \cdot 10^{-5}$\\ 
        \hline
        {\tt France\_Lyon} & 1.25 & 4.89 & 27 $\cdot 10^{-5}$ & $82 \cdot 10^{-5}$\\
        \hline
        {\tt France} & 0.99 & 4.11 & 50 $\cdot 10^{-5}$& $153 \cdot 10^{-5}$\\
        \hline
        \end{tabular}
\end{center}
  \caption{Absolute Errors for the Voltage Magnitude at the Coupling Buses and the Active Power Flow of the Coupling Branches.}
  \label{table:cferrors}
\end{table}

\subsection{Performance of the Learning Models} 

This section compares the model $\mathbb{O}$ that directly
approximates the mapping $\mathcal{O}$ (Equation \ref{original}) with
the proposed two-stage approach $\mathbb{D}$. The results show that,
with a time limit of 90 minutes, $\mathbb{D}$ outperforms $\mathbb{O}$
and is more scalable.  In model $\mathbb{D}$, 30 minutes is allocated
to the first stage, and 60 minutes to the second stage.  The
comparison is performed on the smaller systems, {\tt France\_EHV} and
LYON, which represents the high-voltage French system and the
high-voltage French with a detailed representation of the Lyon
region. Experimental results on the full French system are only given
for model $\mathbb{D}$, since the original model exceeds the capacity
of the GPU memory. The comparison consists of three parts. The first
part reports the accuracy for variables $(\hat{\boldsymbol{v}},
\hat{\boldsymbol{p}}^g)$ (that are directly predicted) and the
indirectly predicted $\hat{\boldsymbol{p}}^f$. The second part
considers constraint violations. The third part discusses how the
predictions can be used to seed an optimization model that restores
feasibility.

\subsubsection{Prediction Accuracy}

\begin{figure*}[t!]
    \centering
    \begin{subfigure}[!t]{0.5\textwidth}
        \centering
        \includegraphics[scale=0.22]{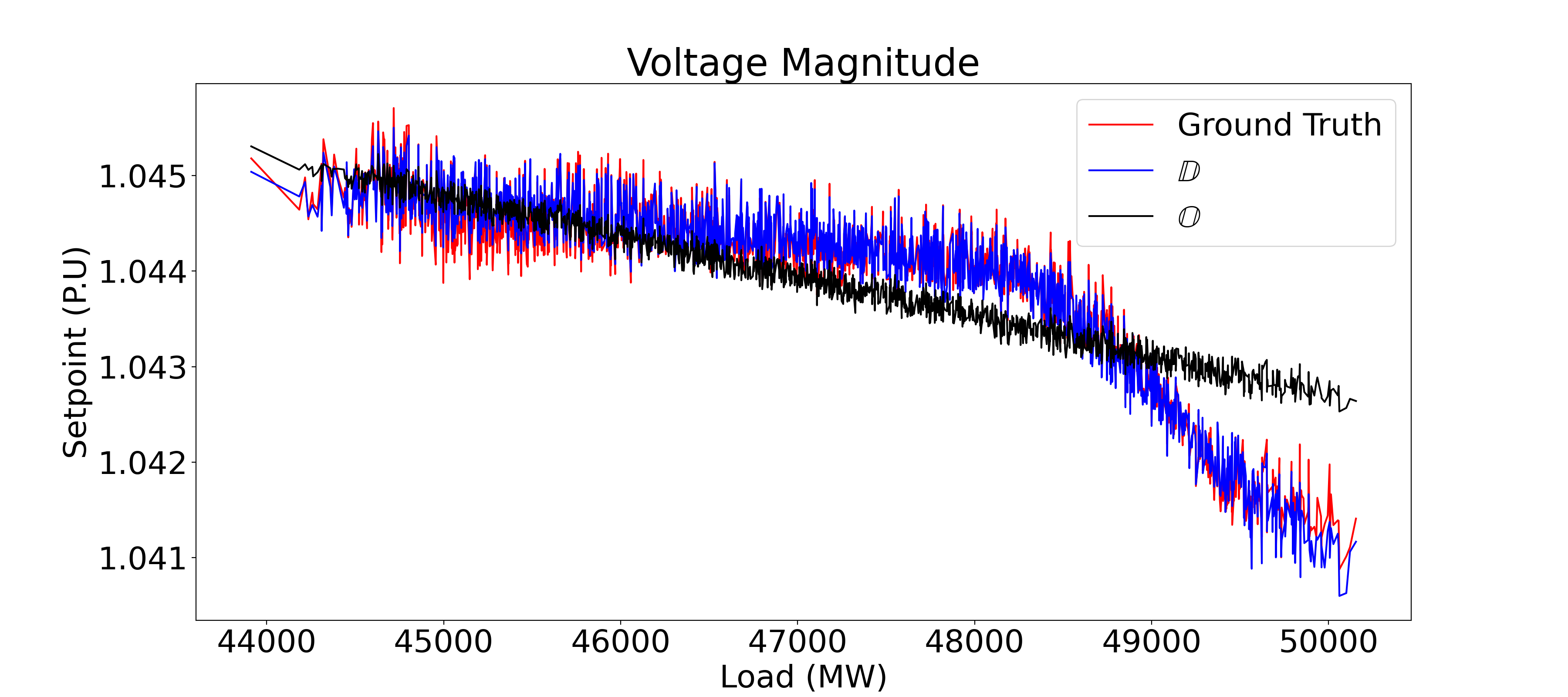}
        \caption{Voltage Magnitude}
    \end{subfigure}%
    \begin{subfigure}[!t]{0.5\textwidth}
        \centering
        \includegraphics[scale=0.22]{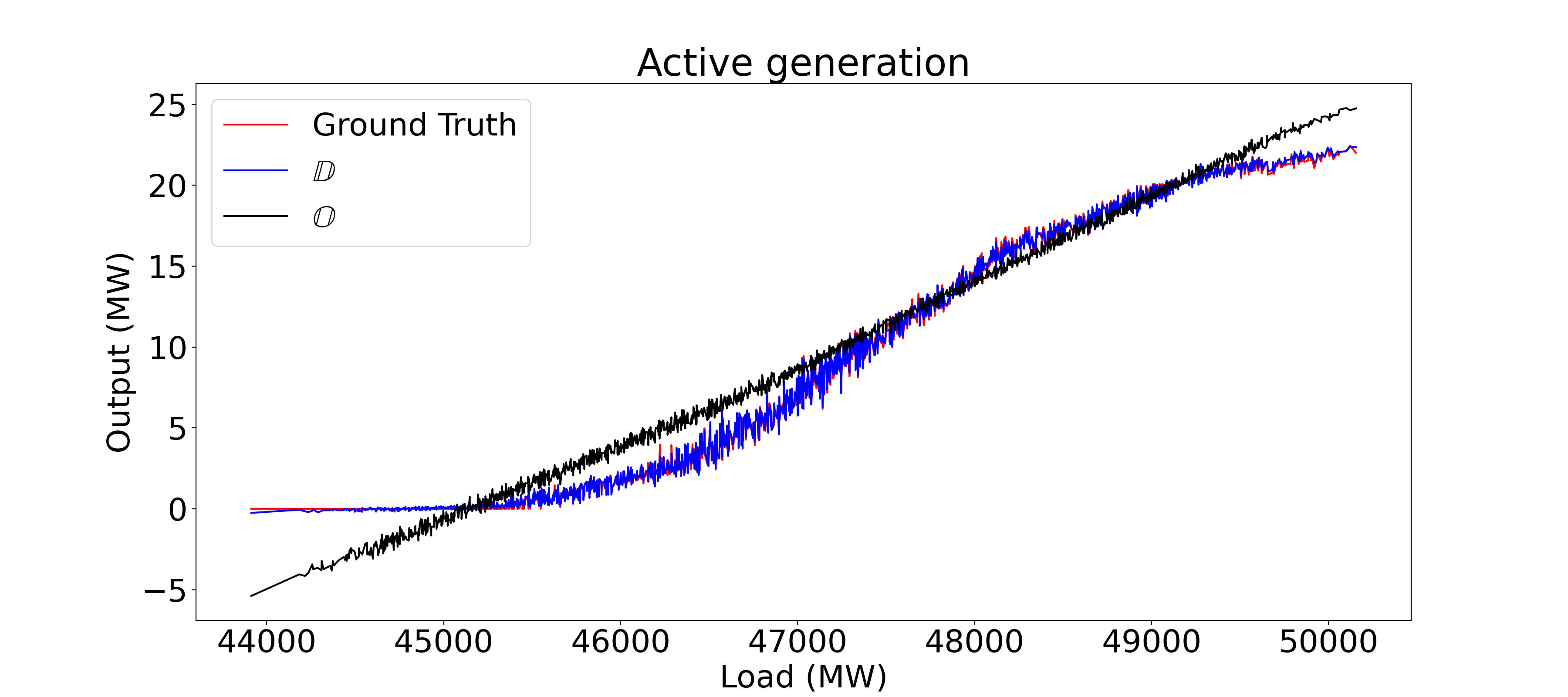}
        \caption{Active power}
    \end{subfigure}
    \caption{Convergence of $\mathbb{O}$ and $\mathbb{D}$ Illustrated for a Bus and Generator.}
    \label{fig:comp}
\end{figure*}

\begin{figure*}[t!]
    \centering
    \begin{subfigure}[!t]{0.5\textwidth}
        \centering
        \includegraphics[scale=0.25]{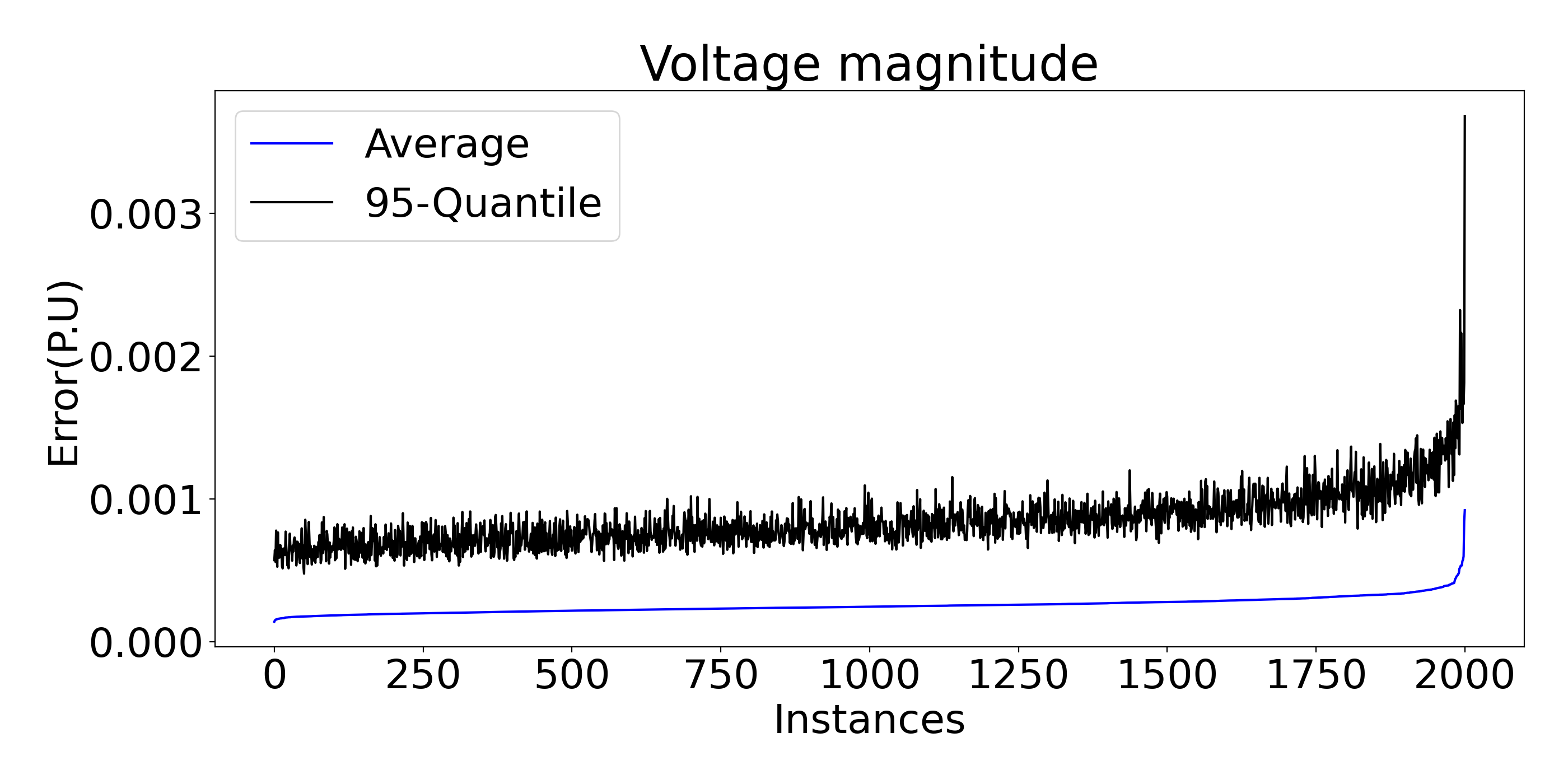}
        \caption{Voltage Magnitude}
    \end{subfigure}%
    ~ 
    \begin{subfigure}[!t]{0.5\textwidth}
        \centering
        \includegraphics[scale=0.25]{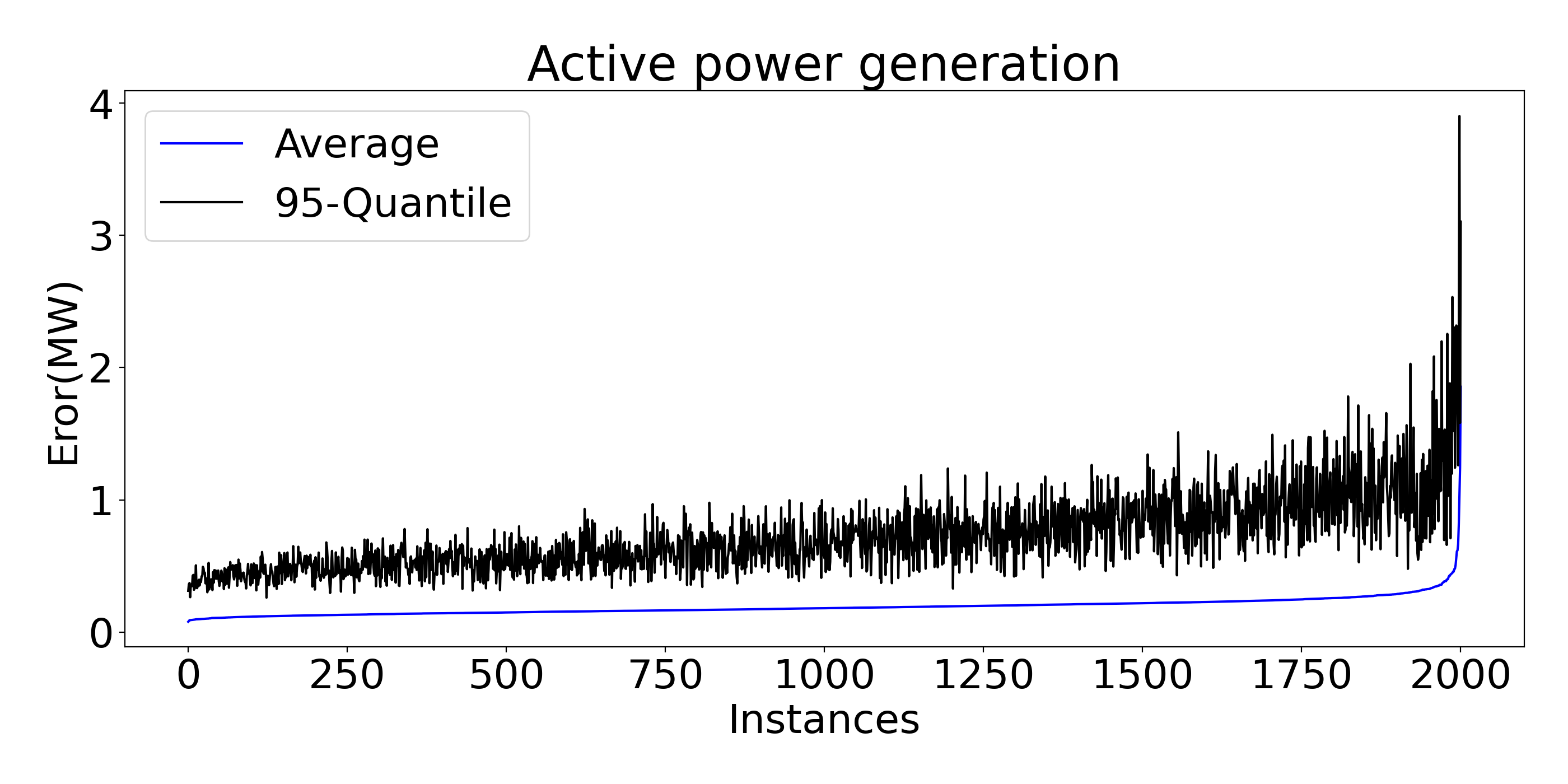}
        \caption{Active power}
    \end{subfigure}
    \caption{Prediction Errors for the {\tt France} System using Model $\mathbb{D}$.}
    \label{fig:predinstances}
\end{figure*}

Figure \ref{fig:comp} illustrate the convergence of the two models for
the predicted variables $\hat{\boldsymbol{v}}, \hat{\boldsymbol{p}}^g$
for a specific bus and generator from the {\tt France\_LYON} test
case. The x-axis corresponds to test instances sorted by increasing
system load. There is significant volatility in the ground truth
values since instances that are close in the x-axis do not necessarily
correspond to similar load vectors. Indeed, a similar overall system
load may exhibit significant regional load differences. The results
demonstrate that, for voltage magnitudes, model $\mathbb{O}$ has
significant errors. The same hold for active power. In constrast,
model $\mathbb{D}$ closely follows the ground truth for voltage
magnitudes and exhibits minor errors for active power predictions. The
difference between the two models is quite striking.

Tables \ref{table:vmerr}, \ref{table:pgerr}, and \ref{table:pferr} summarize the
tprediction errors over all testcases, buses, generators, and lines, as
well the $95\%$ Quantile. The tables omit all power results for
generators that are either off for all instances (due to potentially
high cost) or constantly producing at their respective upper bounds
(due to low cost). For voltage magnitudes, model $\mathbb{D}$ divides
the error in half compared to model $\mathbb{O}$. This difference is
significant for the prediction of the power flows and constraint
violations. Figure \ref{fig:predinstances} demonstrates that model
$\mathbb{D}$ scales to the size of the {\tt France} system and
continues to produce highly accurate predictions. For active power,
model delivers predictions whose errors are an order of magnitude
smaller than those of model $\mathbb{O}$. The average errors are below
$1 \ \mbox{MW}$, which is small compared to the total system load
($\sim$ $50,000 \ \mbox{MW}$). Again, Figure \ref{fig:predinstances}
demonstrates that model $\mathbb{D}$ nicely scales to the {\tt France}
system. The benefits of model $\mathbb{D}$ are abundantly clear for
the power flow predictions $\hat{p}^f$, which are indirectly predicted
as a function of the predictions $\hat{v}$ and $\hat{\theta}$. For
{\tt France\_LYON}, the second largest test case, model $\mathbb{O}$ results
in large errors (up to $50 \ \mbox{MW}$). In contrast, model
$\mathbb{D}$ results in minor errors, with $95$\% of the predictions
having an error of at most $1.04 \ \mbox{MW}$ in the largest
benchmark. Compared to the overall system scale, these errors are
small in percentage. Note that accurate predictions for power flows
are critical for low violation degrees of the AC-OPF constraints.

\begin{table}[!t]
    \begin{center}
    \def\arraystretch{1.2}%
        \begin{tabular}{|c|p{1.2cm} p{1.3cm}|p{1.2cm} p{1.3cm}|} 
        \hline
        & \multicolumn{2}{|c|}{Model $\mathbb{O}$} & \multicolumn{2}{|c|}{Model $\mathbb{D}$} \\
        \hline
        Benchmark & Avg & $95\%$ Quantile & Avg & $95\%$ Quantile\\ [0.5ex] 
        \hline\hline
        {\tt France\_EHV} & $39 \cdot 10^{-5}$ & $125 \cdot 10^{-5}$ & $22 \cdot 10^{-5}$ & $61 \cdot 10^{-5}$\\ 
        \hline
        {\tt France\_Lyon} & $45 \cdot 10^{-5}$ & $127 \cdot 10^{-5}$ & $22 \cdot 10^{-5}$ & $78 \cdot 10^{-5}$\\
        \hline
        {\tt France} & - & - & $25 \cdot 10^{-5}$ & $84 \cdot 10^{-5}$\\
        \hline
        \end{tabular}
\end{center}
  \caption{Prediction Errors (P.U.) for Voltage Magnitudes ($\hat{v}$).}
  \label{table:vmerr}
\end{table}

\begin{table}[!t]
    \begin{center}
    \def\arraystretch{1.2}%
        \begin{tabular}{|c |c c| c c|} 
        \hline
        & \multicolumn{2}{|c|}{Model $\mathbb{O}$} & \multicolumn{2}{|c|}{Model $\mathbb{D}$} \\
        \hline
        Benchmark & Avg & $95\%$ Quantile & Avg & $95\%$ Quantile\\ [0.5ex] 
        \hline\hline
        {\tt France\_EHV} & $8.41$ & $50.82$ & $0.84$ & $3.27$\\ 
        \hline
        {\tt France\_Lyon} & $8.93$ & $47.54$ & $0.30$ & $0.94$\\
        \hline
        {\tt France} & - & - & $0.19$ & $0.70$\\
        \hline
        \end{tabular}
\end{center}
  \caption{Prediction Errors (MW) for Active Power ($\hat{p}^g$).}
  \label{table:pgerr}
\end{table}

\begin{table}[!t]
    \begin{center}
    \def\arraystretch{1.2}%
        \begin{tabular}{|c |c c| c c|} 
        \hline
        & \multicolumn{2}{|c|}{Model $\mathbb{O}$} & \multicolumn{2}{|c|}{Model $\mathbb{D}$} \\
        \hline
        Benchmark & Avg & $95\%$ Quantile & Avg & $95\%$ Quantile\\ [0.5ex] 
        \hline\hline
        {\tt France\_EHV} & $4.53$ & $16.88$ & $2.01$ & $4.20$\\ 
        \hline
        {\tt France\_Lyon} & $8.43$ & $32.20$ & $0.82$ & $1.91$\\
        \hline
        {\tt France} & - & - & $0.45$ & $1.04$\\
        \hline
        \end{tabular}
\end{center}
  \caption{Prediction Errors (MW) for Active Power Flow ($\hat{p}^f$).}
  \label{table:pferr}
\end{table}

\begin{table}[!t]
    \begin{center}
    \def\arraystretch{1.2}%
        \begin{tabular}{|c |c c| c c|} 
        \hline
        & \multicolumn{2}{|c|}{Model $\mathbb{O}$} & \multicolumn{2}{|c|}{Model $\mathbb{D}$} \\
        \hline
        Benchmark & $v$ & $p^g$ & $v$ & $p^g$\\ [0.5ex] 
        \hline\hline
        {\tt France\_EHV} & $99.90$ & $98.46$ & $99.74$ & $99.94$\\ 
        \hline
        {\tt France\_Lyon} & $99.72$ & $99.24$ & $99.91$ & $99.99$\\
        \hline
        {\tt France} & - & - & $99.97$ & $99.99$\\
        \hline
        \end{tabular}
\end{center}
  \caption{Percentage of AC-OPF bound Constraints with Violations under $1 \ \mbox{MW}$ (for $\hat{p}^g$) and under $10^{-4}$ P.U. (for $\hat{v}$)}
  \label{table:sat}
\end{table}

\begin{table}[!t]
    \begin{center}
    \def\arraystretch{1.2}%
        \begin{tabular}{|c |c c| c c|} 
        \hline
        & \multicolumn{2}{|c|}{Model $\mathbb{O}$} & \multicolumn{2}{|c|}{Model $\mathbb{D}$} \\
        \hline
        Benchmark & Avg & $95\%$ Quantile & Avg & $95\%$ Quantile\\ [0.5ex] 
        \hline\hline
        {\tt France\_EHV} & $4.49$ & $16.20$ & $4.82$ & $9.77$\\ 
        \hline
        {\tt France\_Lyon} & $18.65$ & $100.36$ & $1.91$ & $4.67$\\
        \hline
        {\tt France} & - & - & $1.05$ & $2.39$\\
        \hline
        \end{tabular}
\end{center}
  \caption{Violation for the Active Power Balance Constraint (MW).}
  \label{table:pberr}
\end{table}

\subsubsection{Feasibility}

Table \ref{table:sat} reports the constraint violations for the bounds
on active power and voltage magnitude (constraints (\ref{con:2}),
(\ref{con:3a})). Model $\mathbb{D}$ has minor violations for $99.9\%$
of these constraints.  Table \ref{table:pberr} report the violations
of the active flow conservation constraints. Again, model $\mathbb{D}$
has an average $1.05 \ \mbox{MW}$ violations in the {\tt France} test
case, which is insignificant compared to the scale of the system.

\subsubsection{Load-Flow Analysis}

This section shows how model $\mathbb{D}$ can also be used for
applications that require a high-quality {\em feasible} solution to
AC-OPF. It presents an optimization model,
called a Load Flow (L-F) model, for finding the feasible AC-OPF
solution that is closest to the prediction of model $\mathbb{D}$,
i.e.,
\begin{align*}
\min & \quad \lvert \lvert \boldsymbol{p^g} - \boldsymbol{\hat{p}^g} \rvert \rvert_2^2 + \lvert \lvert \boldsymbol{v} - \boldsymbol{\hat{v}} \rvert \rvert_2^2 \tag{$14$} \label{lf}\\
\text{s.t} & \quad \text{(\ref{con:2}) - (\ref{con:6b}}).
\end{align*}

\begin{table}[t!]
    \begin{center}
    \def\arraystretch{1.2}%
        \begin{tabular}{|c |c c| c c|} 
        \hline
        & \multicolumn{2}{|c|}{Model $\mathbb{O}$} & \multicolumn{2}{|c|}{Model $\mathbb{D}$} \\
        \hline
        Benchmark & Avg & Max & Avg & Max \\ [0.5ex] 
        \hline\hline
        {\tt France\_EHV} & $0.036$ & $0.193$ & $0.026$ & $0.119$\\ 
        \hline
        {\tt France\_Lyon} & $0.281$ & $0.987$ & $0.016$ & $0.071$\\
        \hline
        {\tt France} & - & - & $0.012$ & $0.030$\\
        \hline
        \end{tabular}
\end{center}
  \caption{Objective Value Increase of the Load-Flow Solution (in $\%$) Compared to AC-OPF Objective.}
  \label{table:lfobj}
\end{table}

\noindent
Table \ref{table:lfobj} reports the objective increase in the load
flow solution with models $\mathbb{D}$ and $\mathbb{O}$ compared to
the AC-OPF solution, i.e.,
$$
\frac{1}{N_{test}}\sum_{i=1}^{N_{test}} \lvert 1 - \frac{cost_{LF}}{cost_{AC}} \rvert \times 100 \%
$$ where $cost_{LF}$ denotes the load flow cost and $cost_{AC}$
denotes the AC-OPF cost. The load flow based on model $\mathbb{D}$ is
within $< 0.03\%$ on average of the AC-OPF solution and one magnitude
smaller compared to the solutions provided by $\mathbb{O}$ on the {\tt
  France\_Lyon} testcase.

\begin{table}[t!]
    \begin{center}
    \def\arraystretch{1.2}%
        \begin{tabular}{|c |p{0.7cm} p{0.7cm}| p{0.7cm} p{0.7cm}| p{0.7cm} p{0.7cm}|} 
        \hline
        & \multicolumn{2}{|c|}{L-F $\mathbb{O}$} & \multicolumn{2}{|c|}{L-F $\mathbb{D}$} 
        & \multicolumn{2}{|c|}{AC-OPF}\\
        \hline
        Benchmark & Avg & Max & Avg & Max & Avg & Max\\ [0.5ex] 
        \hline\hline
        {\tt France\_EHV} & $7.14$ & $11.72$ & $6.70$ & $10.83$ & $7.10$ & $10.51$ \\ 
        \hline
        {\tt France\_Lyon} & $38.21$ & $186.71$ & $20.08$ & $26.75$ & $75.20$ & $215.47$\\
        \hline
        {\tt France} & - & - & $65.21$ & $130.67$ & $164.07$ & $315.97$\\
        \hline
        \end{tabular}
\end{center}
    \caption{Comparison of the Load Flow and AC-OPF Computation Times (in seconds).}
  \label{table:lftime}
\end{table}

In terms of computational efficiency, model $\mathbb{D}$ delivers a
prediction in a few milliseconds, which makes it sufficient to compare
the optimization results only. Table \ref{table:lftime} compares the
execution times of the load flow and the AC-OPF optimizations. The
results demonstrate that the load flow optimization is significantly
faster compared to the AC-OPF optimization on the largest two
benchmarks. {\em This indicates that a combination of machine learning and
optimization is beneficial when a near-optimal AC-feasible solution to
the OPF is desired.}

\section{Conclusion}

This paper considered the design of a fast and scalable training for a
machine-learning model that predicts AC-OPF solutions. It was
motivated by the facts that (1) topology optimization and the
stochasticity induced in renewable energy may lead to fundamentally
different AC-OPF instances; and (2) existing machine-learning
algorithms for AC-OPF require significant training time of and do not
scale to the size of real transmission systems. The paper proposed a
novel 2-stage approach that exploits a spatial network decomposition.
The power network is viewed as a set of regions, the first stage
learns to predict the flows and voltages on the buses and lines that
are coupling the regions, and the second stage trains, in parallel,
the machine-learning models for each region. Experimental results on
the French transmission system (up to 6,700 buses and 9,000 lines)
demonstrate the potential of the approach. Within a training time of
90 minutes, the approach predicts AC-OPF solutions with very high
fidelity (e.g., an average error of 1 MW for an overall load of 50 GW)
and minor constraint violations, producing significant improvements
over the state-of-the-art.  The results also show that the predictions
can be used to seed a load flow optimization that returns a feasible
solution within $0.03\%$ of the AC-OPF objective, while reducing the
running times by a factor close to 3. Future work will focus on
generalizing the approach to security-constrained OPF, by studying how
to merge the algorithm proposed in \cite{velloso2019exact} to the
AC setting and the proposed 2-stage approach.


\end{document}